\pdfoutput=1

\documentclass[11pt]{article}

\usepackage{acl}

\usepackage{times}
\usepackage{latexsym}
\usepackage{pifont}
\usepackage{booktabs}
\usepackage{xcolor}
\usepackage{array}
\usepackage{multirow}
\usepackage{inconsolata}
\usepackage{amsmath,graphicx}
\usepackage{tabularx}
\usepackage{amssymb}
\usepackage{amsthm}
\usepackage{cleveref}
\usepackage{pifont}
\usepackage{cancel}
\usepackage{lipsum}
\usepackage{fvextra}
\usepackage{booktabs}
\usepackage{algorithm}
\usepackage{algorithmic}
\usepackage{enumitem}
\usepackage{ragged2e}
\usepackage{subcaption}
\usepackage{url}

\usepackage[T1]{fontenc}

\usepackage[utf8]{inputenc}

\usepackage{microtype}

\usepackage{inconsolata}

\usepackage{graphicx}
\usepackage{booktabs}
\usepackage{multirow}
\usepackage[most]{tcolorbox}
\usepackage{titling}

\newcommand{\sdf}{\textsc{SpeechDialogueFactory}}
\newcommand{\sdfs}{\textsc{SDF}}
\newcommand{\sdfen}{\textbf{\textsc{SDF\_en}}}
\newcommand{\sdfzh}{\textbf{\textsc{SDF\_zh}}}
\newcommand{\dt}{\textsc{DailyTalk}}
\newcommand{\cv}{\textsc{CosyVoice2}}
\newcommand{\ftts}{\textsc{Fish-TTS}}

\title{\includegraphics[height=1em]{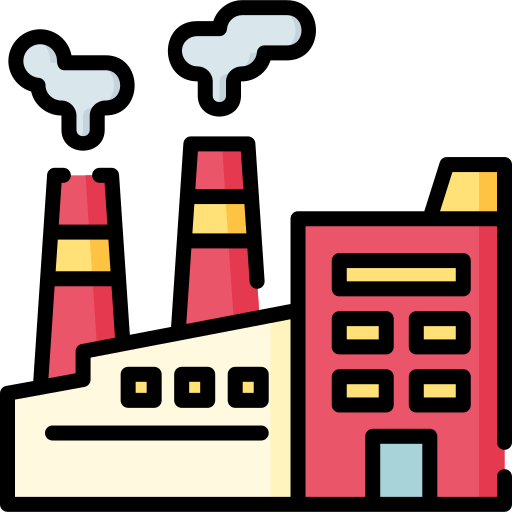} SpeechDialogueFactory: Generating High-Quality Speech Dialogue Data to Accelerate Your Speech-LLM Development}

\author{Minghan Wang\textsuperscript{1}, Ye Bai\textsuperscript{1}, Yuxia Wang\textsuperscript{2}\\ \textbf{Thuy-Trang Vu}\textsuperscript{1},  \textbf{Ehsan Shareghi}\textsuperscript{1}, \textbf{Gholamreza Haffari}\textsuperscript{1} \\
  \textsuperscript{1}Department of Data Science \& AI, Monash University \quad \textsuperscript{2}MBZUAI \\
  \texttt{\{firstname.lastname\}}@monash.edu\\
  \texttt{yuxia.wang}@mbzuai.ac.ae
}

\begin{document}
\maketitle

\begin{abstract}
    High-quality speech dialogue datasets are crucial for Speech-LLM development, yet existing acquisition methods face significant limitations. Human recordings incur high costs and privacy concerns, while synthetic approaches often lack conversational authenticity. To address these challenges, we introduce \textsc{SpeechDialogueFactory}, a production-ready framework for generating natural speech dialogues efficiently. Our solution employs a comprehensive pipeline including metadata generation, dialogue scripting, paralinguistic-enriched utterance simulation, and natural speech synthesis with voice cloning. Additionally, the system provides an interactive UI for detailed sample inspection and a high-throughput batch synthesis mode. Evaluations show that dialogues generated by our system achieve a quality comparable to human recordings while significantly reducing production costs. We release our work as an open-source toolkit\footnote{\url{http://github.com/yuriak/SpeechDialogueFactory}}, alongside example datasets available in English\footnote{\url{https://huggingface.co/datasets/minghanw/sdf\_dataset\_en}} and Chinese\footnote{\url{https://huggingface.co/datasets/minghanw/sdf\_dataset\_zh}}, empowering researchers and developers in Speech-LLM research and development. 
\end{abstract}


\begin{table*}[t]
\centering
\resizebox{\textwidth}{!}{%
\begin{tabular}{@{}lcccccc@{}}
\toprule
Dialogue Generation Framework & \begin{tabular}[c]{@{}c@{}}No Tuning\\ Required\end{tabular} & \begin{tabular}[c]{@{}c@{}}Text \\ Generation\end{tabular} & \begin{tabular}[c]{@{}c@{}}Speech \\ Generation\end{tabular} & \begin{tabular}[c]{@{}c@{}}Emotion \\ Awareness\end{tabular} & \begin{tabular}[c]{@{}c@{}}Multilingual \\ Support\end{tabular} & \begin{tabular}[c]{@{}c@{}}Interactive \\ UI\end{tabular} \\ \midrule
\textsc{PLACES} \cite{chen2023placespromptinglanguagemodels} & \textcolor{green}{\ding{51}} & \textcolor{green}{\ding{51}} & \textcolor{red}{\ding{55}} & \textcolor{red}{\ding{55}} & \textcolor{red}{\ding{55}} & \textcolor{red}{\ding{55}} \\
\textsc{CHATS} \cite{mitsui2023humanlikespokendialoguegeneration} & \textcolor{green}{\ding{51}} & \textcolor{red}{\ding{55}} & \textcolor{green}{\ding{51}} & \textcolor{red}{\ding{55}} & \textcolor{red}{\ding{55}} & \textcolor{red}{\ding{55}} \\
\textsc{Style-Talker} \cite{li2024styletalkerfinetuningaudiolanguage} & \textcolor{red}{\ding{55}} & \textcolor{red}{\ding{55}} & \textcolor{green}{\ding{51}} & \textcolor{green}{\ding{51}} & \textcolor{red}{\ding{55}} & \textcolor{red}{\ding{55}} \\
\textsc{CSR-data} \cite{cornell2024generatingdatatexttospeechlargelanguage} & \textcolor{red}{\ding{55}} & \textcolor{green}{\ding{51}} & \textcolor{green}{\ding{51}} & \textcolor{red}{\ding{55}} & \textcolor{red}{\ding{55}} & \textcolor{red}{\ding{55}} \\
\textsc{PSYDIAL} \cite{han2024psydialpersonalitybasedsyntheticdialogue} & \textcolor{green}{\ding{51}} & \textcolor{green}{\ding{51}} & \textcolor{red}{\ding{55}} & \textcolor{green}{\ding{51}} & \textcolor{red}{\ding{55}} & \textcolor{red}{\ding{55}} \\
\textsc{RefGPT} \cite{yang2023refgptdialoguegenerationgpt} & \textcolor{green}{\ding{51}} & \textcolor{green}{\ding{51}} & \textcolor{red}{\ding{55}} & \textcolor{red}{\ding{55}} & \textcolor{red}{\ding{55}} & \textcolor{red}{\ding{55}} \\ \midrule
\textsc{SpeechDialogueFactory} (ours) & \textcolor{green}{\ding{51}} & \textcolor{green}{\ding{51}} & \textcolor{green}{\ding{51}} & \textcolor{green}{\ding{51}} & \textcolor{green}{\ding{51}} & \textcolor{green}{\ding{51}} \\ \bottomrule
\end{tabular}%
}
\caption{Comparison of representative dialogue generation frameworks from 6 perspectives.}
\label{tab:framework_comparison}
\vspace{-1em}
\end{table*}
\section{Introduction}

The emergence of voice-based AI interfaces has accelerated with recent breakthroughs in large language models (LLMs), establishing Speech-LLMs~\citep{défossez2024moshispeechtextfoundationmodel,chu2023qwenaudioadvancinguniversalaudio,openai2024gpt4ocard} as essential components in next-generation human-computer interaction systems. Developing effective Speech-LLMs involves two crucial phases: cross-modal alignment through joint pretraining of text and speech units \citep{zhang2024omniflattenendtoendgptmodel,xie2024miniomni2opensourcegpt4ovision,zeng2024glm4voiceintelligenthumanlikeendtoend}, followed by specialized training on speech dialogue datasets to develop conversational capabilities \citep{zeng2024glm4voiceintelligenthumanlikeendtoend,wang2024freezeomnismartlowlatency}. High-quality speech dialogue datasets are fundamental to this second phase, requiring both linguistically authentic dialogue content \citep{zhan2023socialdialbenchmarksociallyawaredialogue,zhong2022unifiedmultidimensionalevaluatortext} and naturalistic prosodic features \citep{chen2024voicebenchbenchmarkingllmbasedvoice,maiti2022speechlmscoreevaluatingspeechgeneration}. However, current approaches to acquiring such datasets face significant limitations that hinder progress in the field \citep{défossez2024moshispeechtextfoundationmodel,wang2024freezeomnismartlowlatency}.
Human-recorded dialogues present substantial practical challenges for dataset creation. While professional voice actor recordings can achieve high quality, they involve prohibitive costs that severely limit dataset size and diversity \citep{lee2023dailytalkspokendialoguedataset,zandie2021ryanspeechcorpusconversationaltexttospeech}. Conversely, spontaneous dialogues often contain elements unsuitable for training, including excessive disfluencies and unstructured responses that can negatively impact model performance \citep{yang2022opensourcemagicdataramcrich,barker2018fifthchimespeechseparation,saito2022studiescorpusjapaneseempathetic,li2022enhancingspeakingstylesconversational,liu2024generativeexpressiveconversationalspeech}.

\begin{figure}
    \centering
    \includegraphics[width=1\columnwidth]{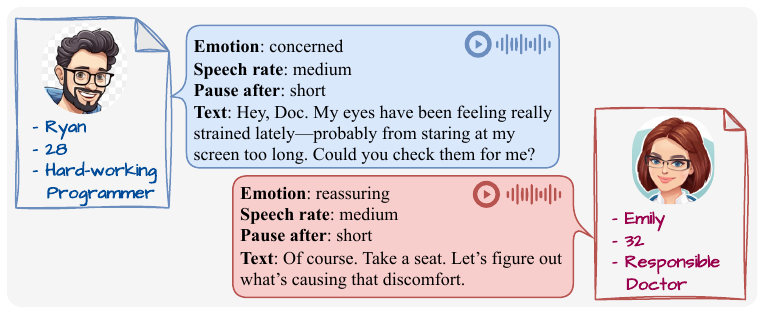}
    \caption{Example dialogue created by \sdf with comprehensive character metadata and paralinguistic annotations.}
    \label{fig:dialogue_example}
    \vspace{-1.7em}
\end{figure}

Therefore, to effectively address the aforementioned challenges in data collection, researchers have increasingly turned to synthetic methods for generating dialogue data. However, current synthetic data generation frameworks still face two critical issues:
(\emph{i}) Insufficient naturalness of generated dialogues. There are distinct differences between conversational and written language styles \citep{zhang2024omniflattenendtoendgptmodel,wang2024freezeomnismartlowlatency}. Dialogues generated by LLMs, which typically excel at producing written text, often lack the natural rhythm, emotional expressiveness, and turn-taking dynamics characteristic of genuine human interactions. (\emph{ii}) Lack of a unified, end-to-end generation framework. As shown in \Cref{tab:framework_comparison}, existing frameworks typically focus separately on either text or speech dialogue generation~\citep{mitsui2023humanlikespokendialoguegeneration, han2024psydialpersonalitybasedsyntheticdialogue}. These methods often require extensive model fine-tuning~\citep{li2024styletalkerfinetuningaudiolanguage} and usually do not incorporate essential features such as emotional expressiveness \cite{chen2023placespromptinglanguagemodels}, multilingual support, or user-friendly interfaces. Consequently, these limitations make it impractical to build large-scale and comprehensive dialogue datasets.

To overcome these limitations, we present \sdf, a systematic framework to generate high-quality speech dialogues with flexible customization and control. Our work makes four key contributions:
\begin{itemize}
    \item \textbf{Quality-Assured Generation Pipeline}: An integrated end-to-end workflow combining structured content creation (metadata, scripting, simulation) with expressive speech synthesis, ensuring both linguistic authenticity and natural prosody through comprehensive paralinguistic annotations.
    \item \textbf{Production-Optimized Implementation}: An efficient system with interactive UI and parallel batch processing capabilities, designed for both exploratory development and large-scale dataset creation.
    \item \textbf{Multilingual Dataset}: Publicly released dialogue datasets in both English and Chinese, along with intermediate records and quality evaluation results to support further research.
\end{itemize}

Our experimental evaluation also confirms that \sdfs~generates dialogues matching the quality of professional recordings, while offering researchers and developers unprecedented flexibility and usability for creating custom speech dialogue datasets tailored to their specific needs.

\section{Related Work}

Currently, numerous frameworks have been proposed in the field of conversational data generation. Among these, some approaches focus primarily on utterance-level generation \cite{sahu2022dataaugmentationintentclassification, chen2022weaklysuperviseddataaugmentation, rosenbaum2022claspfewshotcrosslingualdata, aher2023usinglargelanguagemodels}, while others aim to generate entire conversations. However, there is presently no unified framework capable of comprehensively generating conversations, from textual content to corresponding speech, while simultaneously supporting emotional expressiveness and multilingual capabilities. As summarized in Table \ref{tab:framework_comparison}, frameworks such as \textsc{Places}, \textsc{RefGPT}, and \textsc{psydial} \cite{chen2023placespromptinglanguagemodels, yang2023refgptdialoguegenerationgpt, han2024psydialpersonalitybasedsyntheticdialogue} predominantly concentrate on text-based conversation generation, whereas \textsc{CHATS} and \textsc{Style-Talker} \cite{mitsui2023humanlikespokendialoguegeneration, li2024styletalkerfinetuningaudiolanguage} specialize in speech-based conversation generation. Although \textsc{CSR-data} \cite{cornell2024generatingdatatexttospeechlargelanguage} is capable of generating both textual and speech conversations, it lacks emotional expressiveness and multilingual support. Additionally, none of the aforementioned frameworks provide user interfaces designed for easy interaction by general users.

Furthermore, each of these existing frameworks demonstrates unique strengths. For instance, the \textsc{Places} \cite{chen2023placespromptinglanguagemodels} framework does not require external seed data; instead, it utilizes high-quality expert-written dialogues as prompts, structured according to the desired conversational output, allowing LLMs to directly generate multi-party conversations. \textsc{Style-Talker} \cite{li2024styletalkerfinetuningaudiolanguage} analyzes the acoustic features of the user's input speech to synthesize responses that align closely with the user's speaking style. Additionally, the \textsc{CSR-data} \cite{cornell2024generatingdatatexttospeechlargelanguage} framework uniquely integrates both textual and speech conversation generation, which is relatively rare among current approaches. Researchers leveraging this framework can produce high-quality synthetic dialogue data suitable for fine-tuning automatic speech recognition (ASR) models.

\section{SpeechDialogueFactory}

\begin{figure*}
    \centering
    \includegraphics[width=1.0\textwidth]{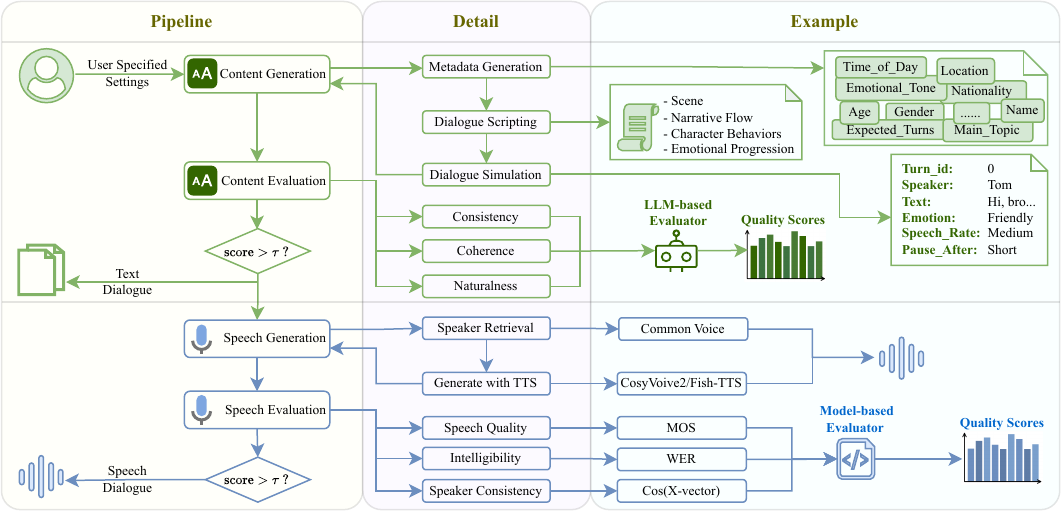}
    \caption{The \sdf~pipeline with integrated quality control. The framework processes user-specified settings through three main stages: content generation (including metadata generation, dialogue scripting, and simulation), speech generation (speaker retrieval and TTS synthesis), and quality evaluation at both text and speech levels to filter low-quality outputs before proceeding to subsequent processing steps.}
    \label{fig:pipeline}
    \vspace{-0.5em}
\end{figure*}

\subsection{Content Generation}
\label{sec:text_synthesis}
Naive approaches to dialogue generation using LLMs (e.g., simple prompts like "Generate a conversation...") prove inadequate for creating high-quality speech dialogue datasets. Our experiments (see \Cref{tab:ablation} baseline) revealed that such methods typically produce limited turns with diminishing coherence and completely lack the paralinguistic features needed for natural speech rendering. To overcome these limitations, \sdfs~implements a structured three-stage pipeline for dialogue content creation that provides both greater control over generation and essential speech synthesis guidance parameters.

\subsubsection{Metadata Generation}
The first stage of our pipeline generates structured metadata that defines the dialogue foundation through three key components: \textbf{dialogue settings} (temporal-spatial context, length parameters), \textbf{character profiles} (demographic information, personality traits, cultural backgrounds), and \textbf{conversation context} (topic progression, emotional arcs). These components are formalized into 26 structured fields within a JSON schema, with generation guided by constrained decoding techniques\cite{willard2023efficientguidedgenerationlarge} to ensure completeness.

To enhance generation diversity while maintaining control, our system implements a \textbf{scenario-seed-based} initialization mechanism operating across five dimensions: dialogue type (e.g., professional discussion, casual conversation), temporal context, spatial setting, cultural background, and language (currently supporting English and Chinese). It also accepts custom prompts for specialized dialogue requirements. Users can directly specify the custom prompt and let the system to complete the scenario seed with appropriate values automatically.

\subsubsection{Dialogue Scripting}
Direct dialogue generation from metadata often produces narrative inconsistencies due to the limited information in structured metadata and the inability to make retroactive corrections. Inspired by \cite{chen2024hollmwoodunleashingcreativitylarge,song2024moviellmenhancinglongvideo}, we implement an intermediate scripting phase that translates metadata into natural language directives. The script serves as a comprehensive blueprint with four components: \textbf{Scene} (environmental context and character relationships), \textbf{Narrative Flow} (dialogue progression through distinct phases), \textbf{Character Behaviors} (interaction patterns and communication styles), and \textbf{Emotional Progression} (affective development with natural transitions). This intermediate representation significantly improves coherence by providing LLMs with comprehensive contextual guidance in a more effective format.

\subsubsection{Dialogue Simulation}

The final content generation stage transforms the metadata and script into actionable dialogue by prompting the LLM to simulate the complete conversation in a single efficient pass. Unlike iterative role-playing approaches that require multiple inference steps, our single-pass implementation significantly reduces computational overhead while maintaining quality.

The system outputs a structured JSON representation of the dialogue, where each utterance is enriched with critical paralinguistic annotations such as emotional state indicators (e.g., "excited," "concerned"), speech rate parameters (slow, medium, fast), and turn-taking dynamics (pause duration between speakers) as shown in \Cref{fig:dialogue_example}.
These annotations serve as control signals for the subsequent speech synthesis stage, enabling more naturalistic rendering of the conversation. The structured format also facilitates systematic quality assessment and potential refinement before advancing to speech generation.

\subsubsection{Content Evaluation}
\label{sec:content_eval}
We implement a comprehensive quality control process that evaluates generated dialogue content before proceeding to resource-intensive speech synthesis. This proactive filtering approach significantly improves overall dataset quality while reducing computational waste. Our system employs a three-dimensional assessment framework that addresses fundamental aspects of dialogue quality: \textbf{consistency, coherence, and naturalness}. Given the inherent subjectivity of dialogue evaluation and absence of ground-truth references \cite{ito2025referencefreeevaluationmetricstext}, we leverage the established LLM-as-a-judge paradigm \cite{zheng2023judgingllmasajudgemtbenchchatbot} to provide reliable and reproducible quality assessments at scale. Through human evaluation, we confirmed the reliability of LLM-based content evaluation (See \Cref{sec:human_eval} for more details).

\paragraph{Consistency}
Our system conducts comprehensive consistency validation across the entire dialogue generation pipeline. We implement an assessment framework with 19 questions (scored 0-100) evaluating three critical dimensions: \textbf{Scenario-Metadata Alignment} (verifying adherence to user parameters, e.g. prompt), \textbf{Metadata Internal Consistency} (examining logical integrity within metadata), and \textbf{Cross-Component Consistency} (ensuring information preservation across pipeline stages). The LLM evaluator analyzes all generated components holistically to produce all consistency scores by filling a checklist.

\paragraph{Coherence}
Our coherence assessment evaluates the dialogue's logical structure and flow at the turn level. The system examines topic relevance, contextual appropriateness of responses, logical flow continuity, absence of contradictions, and overall conversation coherence. Each field receives a 0-100 score from an LLM evaluator, and an overall coherence score is computed by averaging the scores of corresponding aspects for all turns. \sdfs~leverages the complete generation chain (metadata, script, and dialogue) as context, providing richer information for more accurate evaluation.

\paragraph{Naturalness}
The naturalness evaluation measures how closely the generated dialogue mirrors authentic human dialogue patterns. Our framework examines oral style authenticity, appropriate utterance length and rhythm, contextual appropriateness of emotional expressions, consistency between content and emotional markers, and naturalness of vocabulary choices. As with coherence, the LLM evaluator assigns 0-100 scores to each aspect at the turn level, which are then aggregated to produce an overall dialogue naturalness rating. The complete generation history provides essential context for this assessment.

\subsection{Speech Generation}
After generating and validating dialogue content, \sdfs~transforms text into authentic conversational speech through a sophisticated two-stage synthesis pipeline designed for voice diversity and paralinguistic expressiveness.

\subsubsection{Speaker Retrieval}

Our system implements a profile-driven voice matching system that draws from the Common Voice dataset~\cite{ardila2020commonvoicemassivelymultilingualspeech}, selected for its extensive demographic annotations across numerous speakers. The retrieval algorithm maps character attributes from dialogue metadata (age, gender, nationality, etc.) to acoustic characteristics, creating a ranked shortlist of candidate voices for each dialogue participant. This approach ensures acoustic embodiment aligns with narrative characterization, enhancing overall dialogue authenticity.

\subsubsection{Generate with TTS}
We integrate state-of-the-art TTS models, primarily \cv~\cite{du2024cosyvoice2scalablestreaming} and \ftts~\cite{liao2024fishspeechleveraginglargelanguage}, selected for their exceptional prosodic control and consistent voice cloning capabilities. The synthesis process leverages two input streams: \textbf{voice prompts} derived from retrieved speaker samples, and \textbf{natural language control prompts} (only used by \cv) constructed from utterance-specific speech rates and emotional expressions.
Our system enhances conversational realism by implementing variable-length inter-utterance silences based on the dialogue's specified pause durations~\cite{SKANTZE2021101178}, creating natural turn-taking dynamics that avoid the mechanical pacing typical of synthetic dialogue datasets.

\subsection{Speech Evaluation}
\label{sec:acoustic_evaluation}
Complementing our content evaluation process, \sdfs~implements comprehensive speech quality assessment to ensure the acoustic integrity of synthesized dialogues. While content checking focuses on semantic and logical properties, speech quality verification addresses the acoustic manifestation of these dialogues through objective, measurement-based evaluation.

\subsubsection{Speech Quality}
Our system evaluates overall speech naturalness and fidelity using Mean Opinion Score (MOS) metrics implemented through the automated \textsc{UTMOSv2}~\citep{baba2024utmosv2}. This approach provides efficient and consistent quality ratings without human evaluator variability. The framework calculates utterance-level scores for each speech segment, then aggregates these into comprehensive dialogue-level quality metrics that reflect the overall listening experience.

\subsubsection{Intelligibility}
\sdfs~quantifies speech clarity through automated speech recognition (ASR) performance testing. The system transcribes synthesized utterances using \textsc{Whisper-large-v3-turbo}~\cite{radford2022robustspeechrecognitionlargescale} and calculates Word Error Rate (WER) against reference texts for English dialogues, or Character Error Rate (CER) for Chinese content. To isolate TTS quality factors from ASR-specific biases, our implementation normalizes case and excludes punctuation from metric calculations.

\subsubsection{Speaker Consistency}
Voice stability across multiple turns from the same character is critical for dialogue immersion. Our framework analyzes voice consistency by extracting speaker embeddings from each utterance using SpeechBrain's~\cite{speechbrain} pretrained x-vector model. The system then computes cosine similarity between consecutive utterances from the same speaker, flagging potential voice inconsistencies when similarity falls below a configurable threshold (default 0.9).

\subsection{Implementation Details}
\subsubsection{User Interfaces}
\sdfs~offers dual interfaces in Python: a Gradio-based web UI for interactive exploration and a command-line interface for large-scale production.
As shown in \Cref{sec:app_ui} \Cref{fig:ui_interface}, our web interface enables real-time visualization of the generation pipeline with configurable parameters and immediate output inspection with a downloadable packed dialogue file provided after the generation. A configuration tool is also provided in the web UI to help the user prepare commands for batched generation. The command-line interface provides efficient batch processing through simple configuration files and parameter specifications.

\subsubsection{LLM Integration}
The system supports both API-based LLM invocation (connecting to providers like OpenAI) and local inference through VLLM. This dual approach balances convenience for exploratory usage with high-throughput batch acceleration for production-scale generation, allowing seamless transition between development and deployment workflows.

\subsubsection{Parallelization}
To optimize throughput, we implement multi-process parallelization across several generation and evaluation stages, such as TTS, intelligibility, and speech quality evaluation, significantly reducing end-to-end production time for large datasets.

\subsection{Sample Dataset}
\begin{table}[t]
\centering
\resizebox{\columnwidth}{!}{%
\begin{tabular}{@{}l|rr|r@{}}
\toprule
\textbf{Feature} & \multicolumn{1}{c}{\textbf{\textsc{SDF\_en}}} & \multicolumn{1}{c|}{\textbf{\textsc{SDF\_zh}}} & \multicolumn{1}{c}{\textbf{\textsc{DTalk}}} \\ \midrule
\# Turns & 32,021 & 11,312 & 23,773 \\
Total Duration (s) & 437,351 & 86,415 & 78,026 \\
Avg Duration / Turn (s) & 13.658 & 7.639 & 3.282 \\
\# Dialogues & 3,168 & 1,005 & 2,541 \\
Avg Duration / Dialogue (s) & 138.053 & 85.985 & 30.706 \\
Avg Turns / Dialogue & 10.108 & 11.256 & 9.356 \\
\# Speakers & 17 & 35 & 2 \\
\# Topics & 16 & 16 & 10 \\
\# Emotions & 17 & 17 & 7 \\ \bottomrule
\end{tabular}%
}
\caption{Comparative statistics of \dt~(DTalk) and \sdf~(SDF) datasets in English and Chinese.}
\label{tab:dataset_stats}
\vspace{-1em}
\end{table}

\label{sec:sample_dataset}
To make our work more accessible to the community, we generated and released sample datasets in both English (3000+ samples) and Chinese (1000+ samples). Using GPT-4o\footnote{gpt-4o-2024-11-20} and Deepseek-V3~\citep{deepseekai2025deepseekv3technicalreport}, we created 300 custom prompts for each language, spanning diverse temporal-spatial settings, cultural contexts, and speaker demographics (Analysis can be found in \Cref{tab:category_stats} in \Cref{sec:app_data}). For each prompt, we generated multiple dialogues with \textsc{Llama-3.3-70B} for English and \textsc{Qwen-2.5-72B}~\citep{qwen2025qwen25technicalreport} for Chinese, and filtered them with quality evaluation to obtain the final datasets. All intermediate outputs and evaluation results are uploaded to support further research. Detailed statistics can be found in \Cref{tab:dataset_stats}.

\section{System Evaluation}

\begin{table}[t]
\centering
\resizebox{\columnwidth}{!}{%
\begin{tabular}{@{}lccccc@{}}
\toprule
\multicolumn{2}{l|}{\textbf{Dimensions}} & \multicolumn{3}{c|}{\textbf{\textsc{SpeechDialogueFactory}}} & \textbf{\textsc{DailyTalk}} \\ \midrule
\multicolumn{6}{c}{Content Evaluation} \\ \midrule
\multicolumn{2}{l|}{\textbf{LLMs}} & \textbf{\textsc{L3.3-70B}} & \textbf{\textsc{L3.1-8B}} & \multicolumn{1}{c|}{\textbf{\textsc{L3.2-1B}}} &  \\ \midrule
\multicolumn{2}{l|}{\textbf{Consistency} $\uparrow$} & \textbf{96.9} $\pm$ 2.3 & 95.9 $\pm$ 3.5 & \multicolumn{1}{c|}{56.6 $\pm$ 13.1} & - \\
\multicolumn{2}{l|}{\textbf{Coherence} $\uparrow$} & \textbf{99.7} $\pm$ 0.5 & 99.6 $\pm$ 0.7 & \multicolumn{1}{c|}{99.0 $\pm$ 1.3} & 99.2 $\pm$ 1.3 \\
\multicolumn{2}{l|}{\textbf{Naturalness} $\uparrow$} & 91.6 $\pm$ 1.8 & \textbf{92.3} $\pm$ 2.6 & \multicolumn{1}{c|}{90.3 $\pm$ 2.8} & 85.9 $\pm$ 5.0 \\ \midrule
\multicolumn{6}{c}{Speech Evaluation} \\ \midrule
\multicolumn{1}{l|}{\textbf{TTS Models}} & \multicolumn{2}{c}{\textbf{\textsc{CosyVoice2}}} & \multicolumn{2}{c|}{\textbf{\textsc{Fish-TTS}}} &  \\ \midrule
\multicolumn{1}{l|}{\textbf{Speech Quality} $\uparrow$} & \multicolumn{2}{c}{\textbf{3.38} $\pm$ 0.14} & \multicolumn{2}{c|}{2.81 $\pm$ 0.18} & 3.28 $\pm$ 0.21 \\
\multicolumn{1}{l|}{\textbf{Intel.} $\downarrow$} & \multicolumn{2}{c}{2.36 $\pm$ 2.85} & \multicolumn{2}{c|}{\textbf{1.72} $\pm$ 1.66} & 7.05 $\pm$ 6.28 \\
\multicolumn{1}{l|}{\textbf{Spk. Consis.} $\uparrow$} & \multicolumn{2}{c}{\textbf{99.90} $\pm$ 0.17} & \multicolumn{2}{c|}{97.56 $\pm$ 3.40} & - \\
\bottomrule
\end{tabular}%
}
\caption{Content and speech quality evaluation comparing \sdf~with \dt. Bold values indicate best performance. Content evaluation shows results across three Llama3 model scales, while speech evaluation compares two TTS systems against human recordings. \dt~lacks consistency evaluation due to absence of metadata, and speaker consistency metrics due to its fixed two-speaker design. Intel. = Intelligibility (WER), Spk. Consis. = Speaker Consistency.}
\label{tab:overall_evaluation}
\vspace{-1em}
\end{table}
We conducted a comprehensive evaluation on \sdfs~output compared to \dt~\citep{lee2023dailytalkspokendialoguedataset}, a widely-used dataset of human-annotated dialogues recorded by professional voice actors. For fair comparison, we generated 2,541 samples (in English, matching \dt's size) as an experimental set using custom prompts sampled via GPT-4o (with identical topics to \dt). The GPT-4o is also used to evaluate content quality (3 dimensions). Since \dt has no metadata and scripts, we only evaluate its coherence and naturalness solely on the dialogue text. For the speech evaluation, we leverage the methodology described in \S\ref{sec:acoustic_evaluation} for both dataset. To analyze model scaling effects, we compared output from three Llama3~\citep{grattafiori2024llama3herdmodels} variants (3.2-1B, 3.1-8B and 3.3-70B). For speech synthesis evaluation, we tested \cv~\cite{du2024cosyvoice2scalablestreaming} and \ftts~\cite{liao2024fishspeechleveraginglargelanguage} using identical \textsc{Llama3.3-70B} generated content.

\subsection{Quality Benchmarking}
Our evaluation results (\Cref{tab:overall_evaluation}) demonstrate \sdfs's effectiveness across both dimensions. For content, we found that model scale significantly impacts consistency (dramatic improvement from 1B to 8B models), while coherence remains consistently high regardless of scale. All of our models produce more natural dialogues than \dt, with the 8B variant achieving optimal results. For speech synthesis, \cv~outperforms both \ftts~and human recordings on quality metrics, while \ftts~excels in intelligibility. Both systems maintain excellent speaker consistency while offering dramatically greater voice diversity (30+ speakers vs. \dt's 2) without sacrificing quality, confirming our framework's ability to match or exceed human recordings while providing substantially greater scalability.


\subsection{Pipeline Component Analysis}
\begin{table}[t]
\centering
\resizebox{\columnwidth}{!}{%
\begin{tabular}{@{}lccc@{}}
\toprule
\textbf{Settings} & \textbf{Coherence} & \textbf{Naturalness} & \textbf{\# Turns} \\ \midrule
\textbf{Baseline} & 90.25 $\pm$ 9.20 & 84.87 $\pm$ 4.67 & 4.06 $\pm$ 1.34 \\
\textbf{+ Metadata} & 95.68 $\pm$ 4.68 & 87.03 $\pm$ 4.46 & 9.38 $\pm$ 2.01 \\
\textbf{+ Script} & 93.39 $\pm$ 8.43 & 84.06 $\pm$ 4.33 & 11.26 $\pm$ 7.72 \\
\textbf{+ Metadata \& Script} & \textbf{97.39} $\pm$ 3.19 & \textbf{89.01} $\pm$ 3.84 & 11.82 $\pm$ 5.31 \\ \bottomrule
\end{tabular}%
}
\caption{Component contribution analysis comparing four generation approaches: baseline (direct generation), metadata-only, script-only, and complete pipeline. Bold values indicate best performance.}
\label{tab:ablation}
\vspace{-1em}
\end{table}

We systematically evaluated the contribution of each pipeline stage through controlled ablation experiments. Using \textsc{Llama-3.1-8B} for generation and the 70B model for evaluation, we compared four configurations across 300 samples: baseline (direct generation), metadata-only, script-only, and the complete pipeline.
Results in Table \ref{tab:ablation} demonstrate that metadata provides the foundation for quality improvements, increasing coherence by 5.4 points and more than doubling dialogue length. While script-only generation shows limited independent benefits, the full pipeline combining metadata and scripting achieves the highest quality scores across all metrics. These findings confirm the value of our multi-stage approach, with each component contributing to different aspects of dialogue quality.




\section{Conclussion}
We present \sdf, a production-ready system for generating high-quality speech dialogues that addresses critical challenges in Speech-LLM training data creation. Our framework combines a multi-stage content generation pipeline with advanced speech synthesis and comprehensive quality assessment, all integrated into an efficient implementation with LLM inference acceleration, parallelization, flexible deployment options, and user-friendly interfaces. Experimental evaluation confirms our approach produces dialogues matching or exceeding human recordings while offering dramatically greater speaker diversity and cost efficiency. By releasing this as an open-source toolkit and corresponding dataset, we empower researchers to efficiently create customized speech dialogue datasets across domains and languages, with future work focused on multi-party dialogues, enhanced interface capabilities, and optimization for enterprise-scale deployment.
\section*{Limitations}
Our system faces two key limitations: 1) Computational demands for high-quality output, as larger models produce better dialogues but require significantly more resources; 2) Cultural representation biases in the underlying LLM that emerge when prompts lack specific directives, defaulting to overrepresented contexts.

\section*{Ethics Statement}
\sdf~is intended for generating conversational datasets to support Speech-LLM development. Potential misuse exists through creation of harmful or misleading content. We encourage additional safeguards to be applied. Generated content may reflect biases present in LLM training data, requiring careful evaluation before downstream use. This system is meant for creating training data, not for presenting content as authentic human conversations.

\bibliography{custom}

\begin{thebibliography}{43}
\providecommand{\natexlab}[1]{#1}

\bibitem[{Aher et~al.(2023)Aher, Arriaga, and Kalai}]{aher2023usinglargelanguagemodels}
Gati Aher, Rosa~I. Arriaga, and Adam~Tauman Kalai. 2023.
\newblock \href {https://arxiv.org/abs/2208.10264} {Using large language models to simulate multiple humans and replicate human subject studies}.
\newblock \emph{Preprint}, arXiv:2208.10264.

\bibitem[{Ardila et~al.(2020)Ardila, Branson, Davis, Henretty, Kohler, Meyer, Morais, Saunders, Tyers, and Weber}]{ardila2020commonvoicemassivelymultilingualspeech}
Rosana Ardila, Megan Branson, Kelly Davis, Michael Henretty, Michael Kohler, Josh Meyer, Reuben Morais, Lindsay Saunders, Francis~M. Tyers, and Gregor Weber. 2020.
\newblock \href {https://arxiv.org/abs/1912.06670} {Common voice: A massively-multilingual speech corpus}.
\newblock \emph{Preprint}, arXiv:1912.06670.

\bibitem[{Baba et~al.(2024)Baba, Nakata, Saito, and Saruwatari}]{baba2024utmosv2}
Kaito Baba, Wataru Nakata, Yuki Saito, and Hiroshi Saruwatari. 2024.
\newblock The t05 system for the {V}oice{MOS} {C}hallenge 2024: Transfer learning from deep image classifier to naturalness {MOS} prediction of high-quality synthetic speech.
\newblock In \emph{IEEE Spoken Language Technology Workshop (SLT)}.

\bibitem[{Barker et~al.(2018)Barker, Watanabe, Vincent, and Trmal}]{barker2018fifthchimespeechseparation}
Jon Barker, Shinji Watanabe, Emmanuel Vincent, and Jan Trmal. 2018.
\newblock \href {https://arxiv.org/abs/1803.10609} {The fifth 'chime' speech separation and recognition challenge: Dataset, task and baselines}.
\newblock \emph{Preprint}, arXiv:1803.10609.

\bibitem[{Chen et~al.(2024{\natexlab{a}})Chen, Zhu, Yang, Shi, Xi, Zhang, Wang, Pu, Zhang, Yang, and Feng}]{chen2024hollmwoodunleashingcreativitylarge}
Jing Chen, Xinyu Zhu, Cheng Yang, Chufan Shi, Yadong Xi, Yuxiang Zhang, Junjie Wang, Jiashu Pu, Rongsheng Zhang, Yujiu Yang, and Tian Feng. 2024{\natexlab{a}}.
\newblock \href {https://arxiv.org/abs/2406.11683} {Hollmwood: Unleashing the creativity of large language models in screenwriting via role playing}.
\newblock \emph{Preprint}, arXiv:2406.11683.

\bibitem[{Chen et~al.(2023)Chen, Papangelis, Tao, Kim, Rosenbaum, Liu, Yu, and Hakkani-Tur}]{chen2023placespromptinglanguagemodels}
Maximillian Chen, Alexandros Papangelis, Chenyang Tao, Seokhwan Kim, Andy Rosenbaum, Yang Liu, Zhou Yu, and Dilek Hakkani-Tur. 2023.
\newblock \href {https://arxiv.org/abs/2302.03269} {Places: Prompting language models for social conversation synthesis}.
\newblock \emph{Preprint}, arXiv:2302.03269.

\bibitem[{Chen et~al.(2022)Chen, Papangelis, Tao, Rosenbaum, Kim, Liu, Yu, and Hakkani-Tur}]{chen2022weaklysuperviseddataaugmentation}
Maximillian Chen, Alexandros Papangelis, Chenyang Tao, Andy Rosenbaum, Seokhwan Kim, Yang Liu, Zhou Yu, and Dilek Hakkani-Tur. 2022.
\newblock \href {https://arxiv.org/abs/2210.14169} {Weakly supervised data augmentation through prompting for dialogue understanding}.
\newblock \emph{Preprint}, arXiv:2210.14169.

\bibitem[{Chen et~al.(2024{\natexlab{b}})Chen, Yue, Zhang, Gao, Tan, and Li}]{chen2024voicebenchbenchmarkingllmbasedvoice}
Yiming Chen, Xianghu Yue, Chen Zhang, Xiaoxue Gao, Robby~T. Tan, and Haizhou Li. 2024{\natexlab{b}}.
\newblock \href {https://arxiv.org/abs/2410.17196} {Voicebench: Benchmarking llm-based voice assistants}.
\newblock \emph{Preprint}, arXiv:2410.17196.

\bibitem[{Chu et~al.(2023)Chu, Xu, Zhou, Yang, Zhang, Yan, Zhou, and Zhou}]{chu2023qwenaudioadvancinguniversalaudio}
Yunfei Chu, Jin Xu, Xiaohuan Zhou, Qian Yang, Shiliang Zhang, Zhijie Yan, Chang Zhou, and Jingren Zhou. 2023.
\newblock \href {https://arxiv.org/abs/2311.07919} {Qwen-audio: Advancing universal audio understanding via unified large-scale audio-language models}.
\newblock \emph{Preprint}, arXiv:2311.07919.

\bibitem[{Cornell et~al.(2024)Cornell, Darefsky, Duan, and Watanabe}]{cornell2024generatingdatatexttospeechlargelanguage}
Samuele Cornell, Jordan Darefsky, Zhiyao Duan, and Shinji Watanabe. 2024.
\newblock \href {https://arxiv.org/abs/2408.09215} {Generating data with text-to-speech and large-language models for conversational speech recognition}.
\newblock \emph{Preprint}, arXiv:2408.09215.

\bibitem[{DeepSeek-AI et~al.(2025)DeepSeek-AI, Liu, Feng, Xue, Wang, Wu, and et~al.}]{deepseekai2025deepseekv3technicalreport}
DeepSeek-AI, Aixin Liu, Bei Feng, Bing Xue, Bingxuan Wang, Bochao Wu, and Chengda~Lu et~al. 2025.
\newblock \href {https://arxiv.org/abs/2412.19437} {Deepseek-v3 technical report}.
\newblock \emph{Preprint}, arXiv:2412.19437.

\bibitem[{Du et~al.(2024)Du, Wang, Chen, Shi, Lv, Zhao, Gao, Yang, Gao, Wang, Yu, Liu, Sheng, Gu, Deng, Wang, Zhang, Yan, and Zhou}]{du2024cosyvoice2scalablestreaming}
Zhihao Du, Yuxuan Wang, Qian Chen, Xian Shi, Xiang Lv, Tianyu Zhao, Zhifu Gao, Yexin Yang, Changfeng Gao, Hui Wang, Fan Yu, Huadai Liu, Zhengyan Sheng, Yue Gu, Chong Deng, Wen Wang, Shiliang Zhang, Zhijie Yan, and Jingren Zhou. 2024.
\newblock \href {https://arxiv.org/abs/2412.10117} {Cosyvoice 2: Scalable streaming speech synthesis with large language models}.
\newblock \emph{Preprint}, arXiv:2412.10117.

\bibitem[{Défossez et~al.(2024)Défossez, Mazaré, Orsini, Royer, Pérez, Jégou, Grave, and Zeghidour}]{défossez2024moshispeechtextfoundationmodel}
Alexandre Défossez, Laurent Mazaré, Manu Orsini, Amélie Royer, Patrick Pérez, Hervé Jégou, Edouard Grave, and Neil Zeghidour. 2024.
\newblock \href {https://arxiv.org/abs/2410.00037} {Moshi: a speech-text foundation model for real-time dialogue}.
\newblock \emph{Preprint}, arXiv:2410.00037.

\bibitem[{Grattafiori et~al.(2024)Grattafiori, Dubey, and et~al.}]{grattafiori2024llama3herdmodels}
Aaron Grattafiori, Abhimanyu Dubey, and Abhinav~Jauhri et~al. 2024.
\newblock \href {https://arxiv.org/abs/2407.21783} {The llama 3 herd of models}.
\newblock \emph{Preprint}, arXiv:2407.21783.

\bibitem[{Han et~al.(2024)Han, Koh, Seo, Chang, and Sohn}]{han2024psydialpersonalitybasedsyntheticdialogue}
Ji-Eun Han, Jun-Seok Koh, Hyeon-Tae Seo, Du-Seong Chang, and Kyung-Ah Sohn. 2024.
\newblock \href {https://arxiv.org/abs/2404.00930} {Psydial: Personality-based synthetic dialogue generation using large language models}.
\newblock \emph{Preprint}, arXiv:2404.00930.

\bibitem[{Ito et~al.(2025)Ito, van Deemter, and Suzuki}]{ito2025referencefreeevaluationmetricstext}
Takumi Ito, Kees van Deemter, and Jun Suzuki. 2025.
\newblock \href {https://arxiv.org/abs/2501.12011} {Reference-free evaluation metrics for text generation: A survey}.
\newblock \emph{Preprint}, arXiv:2501.12011.

\bibitem[{Lee et~al.(2023)Lee, Park, and Kim}]{lee2023dailytalkspokendialoguedataset}
Keon Lee, Kyumin Park, and Daeyoung Kim. 2023.
\newblock \href {https://arxiv.org/abs/2207.01063} {Dailytalk: Spoken dialogue dataset for conversational text-to-speech}.
\newblock \emph{Preprint}, arXiv:2207.01063.

\bibitem[{Li et~al.(2022)Li, Meng, Li, Wu, Meng, Weng, and Su}]{li2022enhancingspeakingstylesconversational}
Jingbei Li, Yi~Meng, Chenyi Li, Zhiyong Wu, Helen Meng, Chao Weng, and Dan Su. 2022.
\newblock \href {https://arxiv.org/abs/2106.06233} {Enhancing speaking styles in conversational text-to-speech synthesis with graph-based multi-modal context modeling}.
\newblock \emph{Preprint}, arXiv:2106.06233.

\bibitem[{Li et~al.(2024)Li, Jiang, Darefsky, Zhu, and Mesgarani}]{li2024styletalkerfinetuningaudiolanguage}
Yinghao~Aaron Li, Xilin Jiang, Jordan Darefsky, Ge~Zhu, and Nima Mesgarani. 2024.
\newblock \href {https://arxiv.org/abs/2408.11849} {Style-talker: Finetuning audio language model and style-based text-to-speech model for fast spoken dialogue generation}.
\newblock \emph{Preprint}, arXiv:2408.11849.

\bibitem[{Liao et~al.(2024)Liao, Wang, Li, Cheng, Zhang, Zhou, and Xing}]{liao2024fishspeechleveraginglargelanguage}
Shijia Liao, Yuxuan Wang, Tianyu Li, Yifan Cheng, Ruoyi Zhang, Rongzhi Zhou, and Yijin Xing. 2024.
\newblock \href {https://arxiv.org/abs/2411.01156} {Fish-speech: Leveraging large language models for advanced multilingual text-to-speech synthesis}.
\newblock \emph{Preprint}, arXiv:2411.01156.

\bibitem[{Liu et~al.(2024)Liu, Hu, Ren, Yin, and Li}]{liu2024generativeexpressiveconversationalspeech}
Rui Liu, Yifan Hu, Yi~Ren, Xiang Yin, and Haizhou Li. 2024.
\newblock \href {https://arxiv.org/abs/2407.21491} {Generative expressive conversational speech synthesis}.
\newblock \emph{Preprint}, arXiv:2407.21491.

\bibitem[{Maiti et~al.(2022)Maiti, Peng, Saeki, and Watanabe}]{maiti2022speechlmscoreevaluatingspeechgeneration}
Soumi Maiti, Yifan Peng, Takaaki Saeki, and Shinji Watanabe. 2022.
\newblock \href {https://arxiv.org/abs/2212.04559} {Speechlmscore: Evaluating speech generation using speech language model}.
\newblock \emph{Preprint}, arXiv:2212.04559.

\bibitem[{Mitsui et~al.(2023)Mitsui, Hono, and Sawada}]{mitsui2023humanlikespokendialoguegeneration}
Kentaro Mitsui, Yukiya Hono, and Kei Sawada. 2023.
\newblock \href {https://arxiv.org/abs/2310.01088} {Towards human-like spoken dialogue generation between ai agents from written dialogue}.
\newblock \emph{Preprint}, arXiv:2310.01088.

\bibitem[{OpenAI et~al.(2024)OpenAI, :, Hurst, Lerer, Goucher, Perelman, Ramesh, and et~al.}]{openai2024gpt4ocard}
OpenAI, :, Aaron Hurst, Adam Lerer, Adam~P. Goucher, Adam Perelman, Aditya Ramesh, and Aidan~Clark et~al. 2024.
\newblock \href {https://arxiv.org/abs/2410.21276} {Gpt-4o system card}.
\newblock \emph{Preprint}, arXiv:2410.21276.

\bibitem[{Qwen et~al.(2025)Qwen, :, Yang, Yang, Zhang, Hui, Zheng, Yu, Li, Liu, Huang, Wei, Lin, Yang, Tu, Zhang, Yang, Yang, Zhou, Lin, Dang, Lu, Bao, Yang, Yu, Li, Xue, Zhang, Zhu, Men, Lin, Li, Tang, Xia, Ren, Ren, Fan, Su, Zhang, Wan, Liu, Cui, Zhang, and Qiu}]{qwen2025qwen25technicalreport}
Qwen, :, An~Yang, Baosong Yang, Beichen Zhang, Binyuan Hui, Bo~Zheng, Bowen Yu, Chengyuan Li, Dayiheng Liu, Fei Huang, Haoran Wei, Huan Lin, Jian Yang, Jianhong Tu, Jianwei Zhang, Jianxin Yang, Jiaxi Yang, Jingren Zhou, Junyang Lin, Kai Dang, Keming Lu, Keqin Bao, Kexin Yang, Le~Yu, Mei Li, Mingfeng Xue, Pei Zhang, Qin Zhu, Rui Men, Runji Lin, Tianhao Li, Tianyi Tang, Tingyu Xia, Xingzhang Ren, Xuancheng Ren, Yang Fan, Yang Su, Yichang Zhang, Yu~Wan, Yuqiong Liu, Zeyu Cui, Zhenru Zhang, and Zihan Qiu. 2025.
\newblock \href {https://arxiv.org/abs/2412.15115} {Qwen2.5 technical report}.
\newblock \emph{Preprint}, arXiv:2412.15115.

\bibitem[{Radford et~al.(2022)Radford, Kim, Xu, Brockman, McLeavey, and Sutskever}]{radford2022robustspeechrecognitionlargescale}
Alec Radford, Jong~Wook Kim, Tao Xu, Greg Brockman, Christine McLeavey, and Ilya Sutskever. 2022.
\newblock \href {https://arxiv.org/abs/2212.04356} {Robust speech recognition via large-scale weak supervision}.
\newblock \emph{Preprint}, arXiv:2212.04356.

\bibitem[{Ravanelli et~al.(2021)Ravanelli, Parcollet, Plantinga, Rouhe, Cornell, Lugosch, Subakan, Dawalatabad, Heba, Zhong, Chou, Yeh, Fu, Liao, Rastorgueva, Grondin, Aris, Na, Gao, Mori, and Bengio}]{speechbrain}
Mirco Ravanelli, Titouan Parcollet, Peter Plantinga, Aku Rouhe, Samuele Cornell, Loren Lugosch, Cem Subakan, Nauman Dawalatabad, Abdelwahab Heba, Jianyuan Zhong, Ju-Chieh Chou, Sung-Lin Yeh, Szu-Wei Fu, Chien-Feng Liao, Elena Rastorgueva, François Grondin, William Aris, Hwidong Na, Yan Gao, Renato~De Mori, and Yoshua Bengio. 2021.
\newblock \href {https://arxiv.org/abs/2106.04624} {{SpeechBrain}: A general-purpose speech toolkit}.
\newblock \emph{Preprint}, arXiv:2106.04624.
\newblock ArXiv:2106.04624.

\bibitem[{Rosenbaum et~al.(2022)Rosenbaum, Soltan, Hamza, Saffari, Damonte, and Groves}]{rosenbaum2022claspfewshotcrosslingualdata}
Andy Rosenbaum, Saleh Soltan, Wael Hamza, Amir Saffari, Marco Damonte, and Isabel Groves. 2022.
\newblock \href {https://arxiv.org/abs/2210.07074} {Clasp: Few-shot cross-lingual data augmentation for semantic parsing}.
\newblock \emph{Preprint}, arXiv:2210.07074.

\bibitem[{Sahu et~al.(2022)Sahu, Rodriguez, Laradji, Atighehchian, Vazquez, and Bahdanau}]{sahu2022dataaugmentationintentclassification}
Gaurav Sahu, Pau Rodriguez, Issam~H. Laradji, Parmida Atighehchian, David Vazquez, and Dzmitry Bahdanau. 2022.
\newblock \href {https://arxiv.org/abs/2204.01959} {Data augmentation for intent classification with off-the-shelf large language models}.
\newblock \emph{Preprint}, arXiv:2204.01959.

\bibitem[{Saito et~al.(2022)Saito, Nishimura, Takamichi, Tachibana, and Saruwatari}]{saito2022studiescorpusjapaneseempathetic}
Yuki Saito, Yuto Nishimura, Shinnosuke Takamichi, Kentaro Tachibana, and Hiroshi Saruwatari. 2022.
\newblock \href {https://arxiv.org/abs/2203.14757} {Studies: Corpus of japanese empathetic dialogue speech towards friendly voice agent}.
\newblock \emph{Preprint}, arXiv:2203.14757.

\bibitem[{Skantze(2021)}]{SKANTZE2021101178}
Gabriel Skantze. 2021.
\newblock \href {https://doi.org/10.1016/j.csl.2020.101178} {Turn-taking in conversational systems and human-robot interaction: A review}.
\newblock \emph{Computer Speech \& Language}, 67:101178.

\bibitem[{Song et~al.(2024)Song, Wang, Sheng, Zhang, Yu, Fan, and Chen}]{song2024moviellmenhancinglongvideo}
Zhende Song, Chenchen Wang, Jiamu Sheng, Chi Zhang, Gang Yu, Jiayuan Fan, and Tao Chen. 2024.
\newblock \href {https://arxiv.org/abs/2403.01422} {Moviellm: Enhancing long video understanding with ai-generated movies}.
\newblock \emph{Preprint}, arXiv:2403.01422.

\bibitem[{Wang et~al.(2024)Wang, Li, Fu, Shen, Xie, Li, Sun, and Ma}]{wang2024freezeomnismartlowlatency}
Xiong Wang, Yangze Li, Chaoyou Fu, Yunhang Shen, Lei Xie, Ke~Li, Xing Sun, and Long Ma. 2024.
\newblock \href {https://arxiv.org/abs/2411.00774} {Freeze-omni: A smart and low latency speech-to-speech dialogue model with frozen llm}.
\newblock \emph{Preprint}, arXiv:2411.00774.

\bibitem[{Willard and Louf(2023)}]{willard2023efficientguidedgenerationlarge}
Brandon~T. Willard and Rémi Louf. 2023.
\newblock \href {https://arxiv.org/abs/2307.09702} {Efficient guided generation for large language models}.
\newblock \emph{Preprint}, arXiv:2307.09702.

\bibitem[{Xie and Wu(2024)}]{xie2024miniomni2opensourcegpt4ovision}
Zhifei Xie and Changqiao Wu. 2024.
\newblock \href {https://arxiv.org/abs/2410.11190} {Mini-omni2: Towards open-source gpt-4o with vision, speech and duplex capabilities}.
\newblock \emph{Preprint}, arXiv:2410.11190.

\bibitem[{Yang et~al.(2023)Yang, Yuan, Fan, Yang, Wang, Wang, and Zhao}]{yang2023refgptdialoguegenerationgpt}
Dongjie Yang, Ruifeng Yuan, Yuantao Fan, Yifei Yang, Zili Wang, Shusen Wang, and Hai Zhao. 2023.
\newblock \href {https://arxiv.org/abs/2305.14994} {Refgpt: Dialogue generation of gpt, by gpt, and for gpt}.
\newblock \emph{Preprint}, arXiv:2305.14994.

\bibitem[{Yang et~al.(2022)Yang, Chen, Luo, Yang, Ye, Cheng, Xu, Jin, Zhang, Zhang, Xie, and Yan}]{yang2022opensourcemagicdataramcrich}
Zehui Yang, Yifan Chen, Lei Luo, Runyan Yang, Lingxuan Ye, Gaofeng Cheng, Ji~Xu, Yaohui Jin, Qingqing Zhang, Pengyuan Zhang, Lei Xie, and Yonghong Yan. 2022.
\newblock \href {https://arxiv.org/abs/2203.16844} {Open source magicdata-ramc: A rich annotated mandarin conversational(ramc) speech dataset}.
\newblock \emph{Preprint}, arXiv:2203.16844.

\bibitem[{Zandie et~al.(2021)Zandie, Mahoor, Madsen, and Emamian}]{zandie2021ryanspeechcorpusconversationaltexttospeech}
Rohola Zandie, Mohammad~H. Mahoor, Julia Madsen, and Eshrat~S. Emamian. 2021.
\newblock \href {https://arxiv.org/abs/2106.08468} {Ryanspeech: A corpus for conversational text-to-speech synthesis}.
\newblock \emph{Preprint}, arXiv:2106.08468.

\bibitem[{Zeng et~al.(2024)Zeng, Du, Liu, Wang, Jiang, Zhao, Dong, and Tang}]{zeng2024glm4voiceintelligenthumanlikeendtoend}
Aohan Zeng, Zhengxiao Du, Mingdao Liu, Kedong Wang, Shengmin Jiang, Lei Zhao, Yuxiao Dong, and Jie Tang. 2024.
\newblock \href {https://arxiv.org/abs/2412.02612} {Glm-4-voice: Towards intelligent and human-like end-to-end spoken chatbot}.
\newblock \emph{Preprint}, arXiv:2412.02612.

\bibitem[{Zhan et~al.(2023)Zhan, Li, Wang, Luo, Feng, Kang, Hua, Qu, Soon, Sharma, Zukerman, Semnani-Azad, and Haffari}]{zhan2023socialdialbenchmarksociallyawaredialogue}
Haolan Zhan, Zhuang Li, Yufei Wang, Linhao Luo, Tao Feng, Xiaoxi Kang, Yuncheng Hua, Lizhen Qu, Lay-Ki Soon, Suraj Sharma, Ingrid Zukerman, Zhaleh Semnani-Azad, and Gholamreza Haffari. 2023.
\newblock \href {https://arxiv.org/abs/2304.12026} {Socialdial: A benchmark for socially-aware dialogue systems}.
\newblock \emph{Preprint}, arXiv:2304.12026.

\bibitem[{Zhang et~al.(2024)Zhang, Cheng, Deng, Chen, Wang, Zheng, Liu, Yu, and Tan}]{zhang2024omniflattenendtoendgptmodel}
Qinglin Zhang, Luyao Cheng, Chong Deng, Qian Chen, Wen Wang, Siqi Zheng, Jiaqing Liu, Hai Yu, and Chaohong Tan. 2024.
\newblock \href {https://arxiv.org/abs/2410.17799} {Omniflatten: An end-to-end gpt model for seamless voice conversation}.
\newblock \emph{Preprint}, arXiv:2410.17799.

\bibitem[{Zheng et~al.(2023)Zheng, Chiang, Sheng, Zhuang, Wu, Zhuang, Lin, Li, Li, Xing, Zhang, Gonzalez, and Stoica}]{zheng2023judgingllmasajudgemtbenchchatbot}
Lianmin Zheng, Wei-Lin Chiang, Ying Sheng, Siyuan Zhuang, Zhanghao Wu, Yonghao Zhuang, Zi~Lin, Zhuohan Li, Dacheng Li, Eric~P. Xing, Hao Zhang, Joseph~E. Gonzalez, and Ion Stoica. 2023.
\newblock \href {https://arxiv.org/abs/2306.05685} {Judging llm-as-a-judge with mt-bench and chatbot arena}.
\newblock \emph{Preprint}, arXiv:2306.05685.

\bibitem[{Zhong et~al.(2022)Zhong, Liu, Yin, Mao, Jiao, Liu, Zhu, Ji, and Han}]{zhong2022unifiedmultidimensionalevaluatortext}
Ming Zhong, Yang Liu, Da~Yin, Yuning Mao, Yizhu Jiao, Pengfei Liu, Chenguang Zhu, Heng Ji, and Jiawei Han. 2022.
\newblock \href {https://arxiv.org/abs/2210.07197} {Towards a unified multi-dimensional evaluator for text generation}.
\newblock \emph{Preprint}, arXiv:2210.07197.

\end{thebibliography}


\clearpage

\section*{Appendix}
\label{sec:appendix}
\appendix

\section{Dataset Analysis}
Our released datasets, \sdfen~and \sdfzh, substantially outperform the baseline \dt~dataset in scale, diversity, and richness. Table \ref{tab:dataset_stats} shows \sdf's advantages with more dialogues, significantly more turns, longer durations, and dramatically greater speaker diversity. It also offers richer conversational structures with more turns per dialogue and covers a broader range of topics and emotions, as shown in Table \ref{tab:category_stats}.

\label{sec:app_data}
\begin{table}[t]
\centering
\resizebox{\columnwidth}{!}{%
\begin{tabular}{@{}ll|ll@{}}
\toprule
\textbf{Topic Categories} & \multicolumn{1}{c|}{\textbf{\%}} & \textbf{Emotion Categories} & \multicolumn{1}{c}{\textbf{\%}} \\ \midrule
Lifestyle \& SocialMedia & 20.5\% & Enthusiastic \& Passionate & 15.1\% \\
Technology \& SocialTransformation & 15.4\% & Grateful \& Appreciative & 14.0\% \\
Culture \& Traditions & 12.1\% & Curious \& Inquisitive & 13.0\% \\
History \& Archaeology & 9.8\% & Positive \& Optimistic & 11.9\% \\
Environment \& Sustainability & 6.7\% & Reflective \& Philosophical & 10.7\% \\
Health \& MedicalCare & 6.0\% & Confident \& Resolute & 10.2\% \\
Psychology \& PersonalDevelopment & 5.9\% & Supportive \& Encouraging & 5.9\% \\
Family \& InterpersonalRelationships & 4.6\% & Misc & 4.7\% \\
Arts \& Literature & 4.6\% & Cautious \& Conservative & 4.6\% \\
Sci-Fi \& Fantasy & 3.6\% & Anxious \& Nervous & 2.6\% \\
Economics \& Business & 3.2\% & Professional \& Instructional & 2.1\% \\
Food \& CulinaryArts & 3.1\% & Somber \& Sorrowful & 1.4\% \\
Politics \& SocialMovements & 2.6\% & Angry \& Frustrated & 1.1\% \\
Misc & 1.0\% & Humorous \& Amused & 1.0\% \\
Sports \& Entertainment & 0.5\% & Confused \& Uncertain & 0.6\% \\
Crime \& Investigation & 0.4\% & Emotional \& DeeplyMoved & 0.6\% \\
 &  & Creative \& Imaginative & 0.5\% \\ \bottomrule
\end{tabular}%
}
\caption{Distribution of dialogue topics and utterance-level emotions in our released datasets (\S\Cref{sec:sample_dataset}). The percentages are calculated based on the combined statistics (Categories are shared and values are summed up for English and Chinese subsets).}
\label{tab:category_stats}
\end{table}

\section{Human Evaluation}

\begin{table}[t]
\centering
\resizebox{\columnwidth}{!}{%
\begin{tabular}{@{}l|ccc|c@{}}
\toprule
\textbf{Dataset} & \multicolumn{3}{c|}{\textbf{\textsc{SpeechDialogueFactory}}} & \multirow{2}{*}{\textbf{\textsc{DailyTalk}}} \\ \cmidrule(r){1-4}
\textbf{Generator} & \textbf{\textsc{Llama3.3-70B}} & \textbf{\textsc{Llama3.1-8B}} & \textbf{\textsc{Llama3.2-1B}} &  \\ \midrule
\textbf{Consistency} & (69.02 , 63.98) & (82.63 , 72.51) & (94.44 , 93.96) & \textbf{-} \\
\textbf{Coherence} & (71.15 , 71.22) & (85.08 , 77.48) & (83.44 , 81.50) & (86.33 , 82.00) \\
\textbf{Naturalness} & (40.23 , 39.79) & (56.66 , 53.34) & (61.49 , 51.18) & (79.45 , 79.23) \\ \bottomrule
\end{tabular}%
}
\caption{Correlation between human and LLM evaluations across different models and datasets. Values shown as (Pearson correlation, Spearman correlation) demonstrate alignment between human and automated quality assessments. Higher values indicate stronger agreement. \dt~lacks consistency evaluation due to the absence of metadata.}
\label{tab:human_eval}
\vspace{-1em}
\end{table}
\label{sec:human_eval}
To validate our LLM evaluator, we conducted human evaluation with three proficient English speakers rating 100 samples from each model setting and \dt. \Cref{tab:human_eval} shows strong correlations between human and LLM judgments across settings, with particularly high agreement on lower-quality samples (Llama3.2-1B). This confirms LLMs can effectively differentiate dialogue quality in alignment with human judgment.

\section{User Interface Screenshot}
\label{sec:app_ui}
\begin{figure*}[t]
    \centering
    \includegraphics[width=1\textwidth]{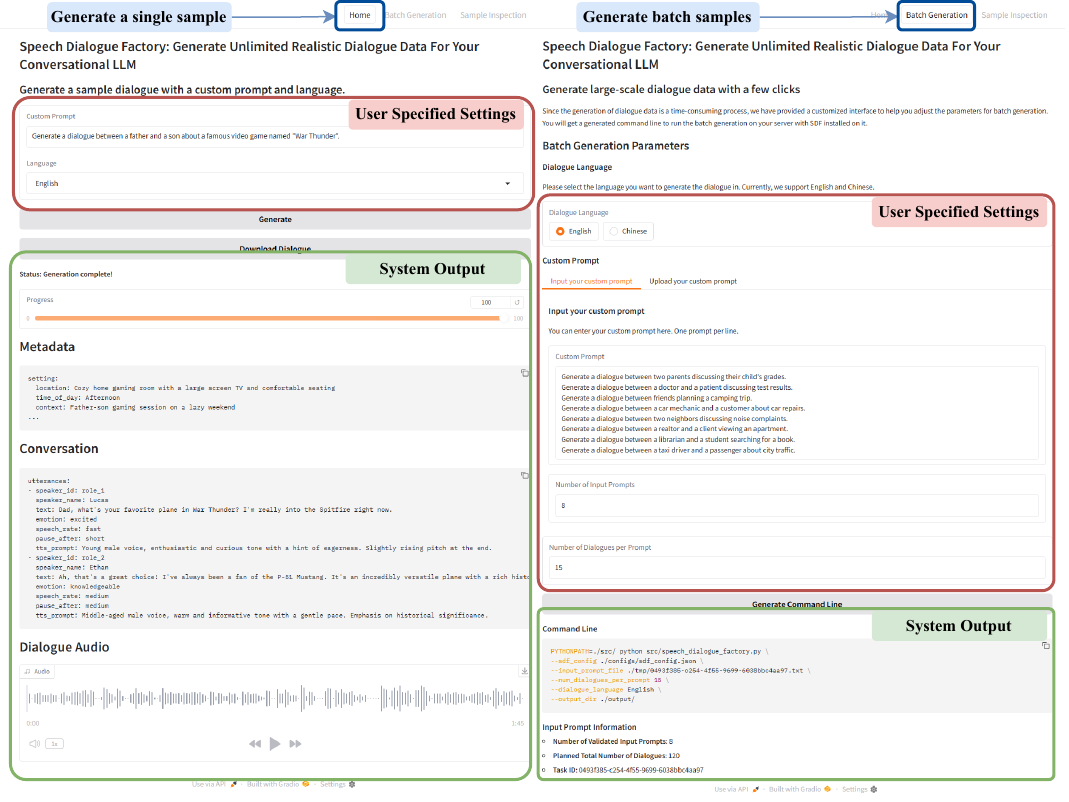}
    \caption{User Interface of \sdf. The main interface of \sdf~consists of 3 tabs: Single Sample Generation (<- left), Batch Samples Generation (-> right), and Sample Inspection. In the \textbf{Single Sample Generation tab} tab (<- left), the System Output section (highlighted in \textcolor{teal}{green}) displays the complete results, including intermediate outputs such as Metadata, Scripts, Dialogue Audio and Quality Scores (note: screenshot content is simplified for clarity). In the \textbf{Batch Samples Generation} tab (-> right), the System Output (highlighted in \textcolor{teal}{green}) provides a pre-generated command line, allowing users to easily copy and paste it into their terminal to initiate batch dialogue generation. The purpose of the \textbf{Sample Inspection} tab is to provide a convenient way to inspect individual dialogue samples generated using the batch method. The display is similar to the one described in the Single Sample Generation tab; thus, we will not present it separately here.}
    \label{fig:ui_interface}
\end{figure*}
\Cref{fig:ui_interface} presents the two major web pages of \sdf, corresponding to the interactive single dialogue generation mode and batched generation model.

\end{document}